\documentclass[twoside,11pt]{article}
\usepackage{jair, theapa, rawfonts}

\usepackage{latexsym}

\usepackage{graphicx}\graphicspath{ {images/} }

\usepackage[round]{natbib}

\usepackage{booktabs}
\usepackage{multirow}

\usepackage{url}
\usepackage[breaklinks]{hyperref}
\hypersetup{breaklinks=true}
\usepackage{xspace}
\usepackage{xcolor}
\usepackage{enumerate}
\usepackage{amsmath}

\newcommand{\SiamL}{\textsc{Siam$_{L_1}$}\xspace}
\newcommand{\SiamCOS}{\textsc{Siam$_{\mbox{cos}}$}\xspace}
\newcommand{\KnownAuth}{\textsc{KnownAuth}\xspace}
\newcommand{\OneShot}{\textsc{OneShot}\xspace}

\newcommand{\Koppel}{\textsc{Koppel}\xspace}
\newcommand{\CNN}{\textsc{cnn}\xspace}

\newcommand{\FFa}{\textsc{ff-100}\xspace}
\newcommand{\FFb}{\textsc{ff-1K}\xspace}
\newcommand{\FFc}{\textsc{ff-5K}\xspace}
\newcommand{\FFd}{\textsc{ff-10K}\xspace}
\newcommand{\FF}{\textsc{ff}\xspace}
\newcommand{\Blog}{\textsc{bl-2K}\xspace}
\newcommand{\PanF}{\textsc{pan15}\xspace}
\newcommand{\PanT}{\textsc{pan20}\xspace}

\newcommand{\mycitet}[1]{\citet{#1}}
\newcommand{\mycitep}[1]{\citep{#1}}

\DeclareMathOperator{\cossim}{cossim}

\ShortHeadings{Siamese Networks for Large-Scale Author Identification}
{Saedi \& Dras}
\firstpageno{1}

\begin{document}

\title{Siamese Networks for Large-Scale Author Identification}

\author{\name Chakaveh Saedi \email chakaveh.saedi@hdr.mq.edu.au \\
       \name Mark Dras \email mark.dras@mq.edu.au \\
       \addr Macquarie University, Department of Computing, Sydney, Australia}


\maketitle




\begin{abstract}
Authorship attribution is the process of identifying the author of a text. Approaches to tackling it have been conventionally divided into classification-based ones, which work well for small numbers of candidate authors, and similarity-based methods, which are applicable for larger numbers of authors or for authors beyond the training set; these existing similarity-based methods have only embodied static notions of similarity.  Deep learning methods, which blur the boundaries between classification-based and similarity-based approaches, are promising in terms of ability to learn a notion of similarity, but have previously only been used in a conventional small-closed-class classification setup.  

Siamese networks have been used to develop learned notions of similarity in one-shot image tasks, and also for tasks of mostly semantic relatedness in NLP.  We examine their application to the stylistic task of authorship attribution on datasets with large numbers of authors, looking at multiple energy functions and neural network architectures, and show that they can substantially outperform previous approaches.  
\end{abstract}

\section{Introduction}
\label{sec:introduction}


Apart from early statistical methods, computational approaches to authorship attribution have conventionally been divided into classification-based and similarity-based \mycitep{stamatatos2009survey}.  In the classification-based paradigm, candidate authors are identified, and texts known to be from them are used to construct a training set; a standard supervised machine learning approach is then typically applied to construct a classifier, with methods now including deep learning \mycitep{ruder2016character}.  This is the more common approach to authorship identification, which has successfully tackled basic versions of the problem, mostly with small numbers of authors, where these authors form a closed set and where they are known in advance.  \mycitet{koppel2011authorship} describe this as the ``vanilla'' version of the problem that is not the typical case in the real world; rather, in real world problems there may be perhaps thousands of candidate authors, the author may not in fact be one of the candidates, and the known text from authors may be limited.

\mycitet{koppel2011authorship} argued that \textit{similarity-based} approaches are better suited to large numbers of candidate authors.  In the similarity-based paradigm, some metric is used to measure the distance between two texts, and an unknown text is attributed to the author of the known one(s) it is closest to, making this one of a class of nearest neighbour approaches \mycitep{hastie-etal:2009}.  For authorship attribution, there have been fewer of these approaches than of the classification-based ones, with noteworthy ones including the Writeprints method of \mycitet{abbasi-chen:2008:TOIS} and the method of \mycitet{koppel2011authorship}.  The empirical success of this latter method has been demonstrated in a set of authorship shared tasks over a number of years, organised as part of the PAN framework of shared tasks on digital forensics and stylometry:\footnote{\url{https://pan.webis.de/}} for example, it formed the core of two of the winners of PAN authorship shared tasks
\mycitep{seidman:2013:PAN,khonji-iraqi:2014:PAN}, and has been a standard inference attacker for the PAN shared task on authorship obfuscation, where the goal is to conceal authorship from methods that attempt to detect it.  
\mycitet{koppel2011authorship} also note that reducing authorship attribution to instances of the binary \textit{authorship verification} problem --- determining if a given document is by a particular author or not --- permits authorship attribution in cases where the author is not one of the known candidates, and is more naturally suited to similarity-based models.  
However, most existing methods have used only a static notion of similarity over fixed features, rather than a learned one.

Subsequent to much of this earlier work, deep learning has led to major changes in NLP, both in terms of learning more accurate models for a range of tasks, but also in blurring the distinction between classification-based and similarity-based approaches: the learned representations can be applied to predicting classes or to determining similarity of data items.  A useful distinction to draw is between closed set and open set recognition \mycitep{geng-etal:2021:TPAMI}: whether models apply to knowledge of the world at training time, as in the conventional classification-based authorship framework, or beyond that, as in \mycitet{koppel2011authorship}'s advocacy for approaching the task via similarity.

There are tasks within NLP that have successfully used deep learning for applying a learned notion of similarity, but these have been applied to learning similarity between semantic aspects of texts, rather than looking at stylistic similarity.  For example, tasks like QA and image captioning have been tackled using Deep Semantic Similarity Models (DSSM), e.g. in the work of \mycitet{yih-etal:2014:ACL} and \mycitet{fang-etal:2015:CVPR}, respectively, in mapping between the semantics of image and corresponding text; other tasks like duplicate question detection \mycitep{rodrigues2017ways} --- identifying questions with the same semantics --- and semantic composition \mycitep{cheng-kartsaklis:2015:EMNLP} have similarly been approached with semantic similarity models.  Approaches like these might also apply to stylistic similarity, with the right feature space, but it is an open question as to their application beyond semantics \mycitep{gerz:2020}. 

A task that has parallels to our own comes from image processing: building on the original use of Siamese networks for signature verification by \mycitet{bromley1994signature}, \mycitet{koch2015siamese} use deep Siamese networks to learn a notion of similarity between images, where the generality of this notion is evaluated via one-shot recognition.  This style of Siamese network has been adapted for a range of semantics-based NLP tasks, such as for sentence similarity by \mycitet{mueller-thyagarajah:2016:AAAI} and for tasks like paraphrase identification by \mycitet{yin-etal:2016:TACL}.  A proposal to use them for the stylistic task of author identification came from \mycitet{dwyer:2017}, but while the idea was appealing, results in that work were not positive, and performed in some cases worse than the baseline.  More recently \mycitet{boenninghoff-etal:2019:ICASSP} applied Siamese networks to short social media texts on a relatively small PAN dataset, producing some positive results.  Contemporaneously with the present work, \mycitet{DBLP:conf/clef/Araujo-PinoGP20} and \mycitet{DBLP:conf/clef/BoenninghoffRNK20} also applied Siamese networks in a closed-set PAN context.

We define a range of Siamese-based architectures for authorship attribution on the sorts of texts standardly used in this task and across large numbers of authors, and evaluate them in both known-author (closed set) and one-shot learning contexts (open set); as part of this, we examine the effect of choice of energy function, sub-network structure and text representation.  We show that they can outperform both a strong classification-based baseline and, in one-shot contexts, the key conventional similarity-based method of \mycitet{koppel2011authorship}, on datasets with large numbers of authors.  We also find clear preferences for choice of sub-network type and text representation.

\section{Related Work}
\label{sec:rel-work}

\subsection{Authorship Identification}

\mycitet{stamatatos2009survey} surveyed approaches up until 2009: we noted in \S\ref{sec:introduction} the division into classification-based and similarity-based approaches, the latter of which is better suited to large numbers of authors.  A key work following the survey was the similarity-based approach of \mycitet{koppel2011authorship}.  The method represents texts by vectors of space-free character 4-grams, and then repeatedly samples features from these vectors and takes the cosine similarity between the vectors consisting of these sampled features; \mycitet{koppel-winter:2014:JASIST} later found that the Ruzicka metric produced better results.  Like the majority of work in authorship identification, these are applied to longer texts; in this specific instance, to blog posts taken from 10,000 authors.\footnote{Taken from \url{blogger.com}.}  Given the successful application to a very large number of authors, we use this as a baseline method in this paper.

Much work on authorship identification since then has appeared in PAN shared tasks: 
the years with attribution setups like this paper were 2011, 2012 and 2018, while the years 2013--2015 considered a verification setup instead.
The attribution tasks have required choosing among small numbers of authors, e.g. 3 for 2012 \mycitep{juola2012overview} up to 20 for 2018 \mycitep{kestemont-etal:2018:PAN}.  For the most part systems in these tasks use conventional machine learning (i.e. not deep learning): the 2018 winner used an ensemble classifier \mycitep{custodio2018each} and the runner-up a linear SVM \mycitep{murauer2018dynamic}.  As noted in \S\ref{sec:introduction}, two earlier winners in verification setups \mycitep{seidman:2013:PAN,khonji-iraqi:2014:PAN} were based on the similarity approach of \mycitet{koppel2011authorship}, which we used as a baseline.  Another exception to conventional machine learning was the 2015 winner, \mycitet{bagnall2015author}, using an RNN-based classifier with shared state but different softmax layer for each author: the architecture is not generally applicable.



The PAN author attribution task in 2019 changed focus from the previous setups and the setup of the present paper, to focus on cross-domain texts; the task overview \mycitep{kestemont2019overview} noted that no deep learning approaches were used there (because of poor performance in the 2018 task, likely due to the small data setup), and participating systems were typically standard ensembles of conventional classifiers.
After this digression, PAN in 2020 returned to closed-set author verification, with plans for open-set verification in 2021 and a `surprise task' in 2022. PAN 2020 differed from previous years with the use of a much larger dataset to support more data-hungry approaches like deep learning \mycitep{Kestemont2020OverviewOT}.

Systems participating in the PAN 2020 verification task were developed contemporaneously with the present work.  Approaches included neural networks \mycitep{DBLP:conf/clef/Araujo-PinoGP20,DBLP:conf/clef/BoenninghoffRNK20,DBLP:conf/clef/OrdonezSC20}, statistical and regression models \mycitep{DBLP:conf/clef/Kipnis20,DBLP:conf/clef/WeerasingheG20}, and comparisons on the basis of specific predefined or extracted features and thresholds \mycitep{DBLP:conf/clef/HalvaniGR20,DBLP:conf/clef/Gagala20,DBLP:conf/clef/Ikae20}.  We discuss the relationship of the two closest of these to our work in \S\ref{sec:rel-work-siamese}.

Outside the PAN framework, 
some work is specific to certain authorship contexts and not purely stylistic: e.g. \mycitet{chen2017task} and \mycitet{zhang2018camel} on scientific authorship, incorporating publication content and references.  Other work uses additional features that are restricted to specific contexts, such as the work by \mycitet{hou-huang:2020:JNLE} that incorporates tone in stylometric analysis for Mandarin Chinese.
Notable  
work on purely stylistic authorship identification with standardly used features, as in this paper, included the use of LDA by \mycitet{seroussi2011authorship}, both within an SVM and using Hellinger distance, to handle large numbers of authors; this was extended in \mycitet{seroussi-etal:2014:CL}.  \mycitet{mohsen2016author} used feature extraction via a stack denoising auto encoder and then classification via SVM.  
\mycitet{ruder2016character} proposed a CNN classification model which outperformed \mycitet{seroussi2011authorship} and various other conventional machine learning approaches on up to 50 authors across a range of datasets; given the set of comparators and the relatively large number of authors used for a classification approach, we use it in this paper as another baseline.

\begin{figure}
  \centering
  \includegraphics[width=60mm]{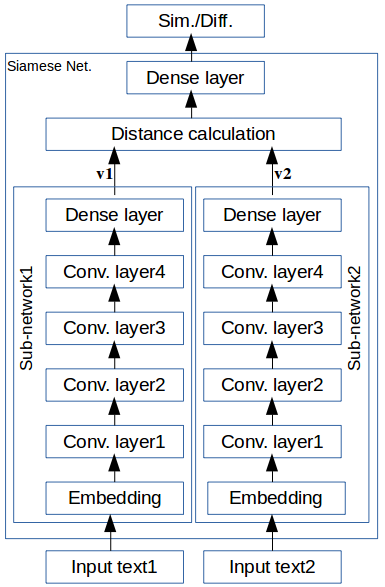}
  \caption{\vspace{-0.75cm}Siamese network architecture.}
  \label{fig:siamese-net}
\vspace{0.5cm}
\end{figure} 

\subsection{Siamese Architectures}
\label{sec:rel-work-siamese}

Siamese networks were first used for verifying signatures, by framing it as an image matching problem \mycitep{bromley1994signature}.  The key features of the Siamese network were that it consisted of twin sub-networks, linked together by an energy function (Fig~\ref{fig:siamese-net}).  The weights on the sub-networks are tied, so that the sub-networks are always identical: inputs are then mapped into the same space, and the energy function represents some notion of distance between them.
%
Siamese networks were updated for deep learning by \mycitet{koch2015siamese} for the task of general image recognition.  The sub-networks were convolutional neural networks (CNNs), and to the outputs of the final layers of these CNNs the weighted $L_1$ distance was calculated and a sigmoid activation applied; a cross-entropy objective was then used in training.

This idea of deep learning-based Siamese networks has been adapted from \mycitet{koch2015siamese} for a number of tasks in NLP.  In most cases, these Siamese networks have been applied to tasks at the level of sentences or below.
\mycitet{mueller-thyagarajah:2016:AAAI} applied a Siamese network structured like that of \mycitet{koch2015siamese} to the problem of sentence similarity using the SICK dataset, where pairs of sentences have been assigned similarity scores derived from human judgements.  Their architecture similarly used $L_1$ (Manhattan) distance, but the sub-networks were LSTMs.  At around the same time, \mycitet{yin-etal:2016:TACL} defined an attention-based model motivated by the Siamese architecture of \mycitet{bromley1994signature}, and applied it to the tasks of answer selection, paraphrase identification and textual entailment.  Much subsequent work has been similar in terms of applications: to answer selection or question answering \mycitep{das-etal:2016:ACL,tay-etal:2018:WSDM,hu:2018:EMNLP,lai-etal:2018:COLING}, sentence similarity \mycitep{reimers-gurevych:2019:EMNLP}, job title normalisation \mycitep{neculoiu-etal:2016},
matching e-commerce items \mycitep{shah-etal:2018:NAACL}, learning argumentation \mycitep{joshi-etal:2018:SemEval,gleize-etal:2019:ACL}, and detecting funnier tweets \mycitep{baziotis-etal:2017:SemEval}.
In some cases Siamese networks have been applied to word-based rather than sentence-based problems, such as identifying cognates \mycitep{rama:2016:COLING} or antonyms \mycitep{etcheverry-wonsever:2019:ACL}.  In other cases the application of Siamese networks is secondary to the main task, such as relation extraction \mycitep{rossiello-etal:2019:NAACL} or supervised topic modelling \mycitep{huang-etal:2018:EMNLP}.  The most typical configuration is to use some kind of RNN as sub-network and cosine similarity for the energy function; but CNNs are also used in sub-networks, and for energy functions $L_1$ and $L_2$ are also used, along with some less common alternatives such as a hyperbolic distance function \mycitep{tay-etal:2018:WSDM} or one based on LSTM-based importance weighting \mycitep{hu:2018:EMNLP}.

There have been two attempts before this work to use Siamese networks for author identification, and two contemporaneous with it.  The first was \mycitet{dwyer:2017}, who observed that within NLP they have been used only for short texts rather than the longer sort standardly used for author identification, which is supported by the above summary of applications of Siamese networks to NLP.  For sub-networks, \mycitet{dwyer:2017} used fully connected networks, and $L_2$ as the energy function.  Experimentally, on data from the PAN 2014 and 2015 tasks, he found that results were fairly poor, in some cases worse than a random baseline.  More recently \mycitet{boenninghoff-etal:2019:ICASSP} applied Siamese networks to short social media texts, using an architecture with LSTMs as sub-networks and an energy function based on Euclidean distance.  This was applied to the relatively small PAN 2016 dataset, and produced some positive results.
In the author verification context of PAN 2020, the winning system of \mycitet{DBLP:conf/clef/BoenninghoffRNK20} used a Siamese network with LSTMs as sub-networks and a Probabilistic Linear Discriminant Analysis layer and Linguistic Embedding Vectors (LEV) to perform Bayes factor scoring for the verification task.  The other Siamese network, of \mycitet{DBLP:conf/clef/Araujo-PinoGP20}, consisted of residual sub-networks with densely connected components, and an $L_1$ energy function.




\section{The Model}
\label{sec:model}

Our architecture follows the basic structure of \mycitet{koch2015siamese}, and of \mycitet{mueller-thyagarajah:2016:AAAI} for sentence similarity; these both used $L_1$ distance for the energy function, although \mycitet{koch2015siamese} used CNNs for the sub-networks while \mycitet{mueller-thyagarajah:2016:AAAI} used LSTMs.  The goal of our network is to produce similarity scores for text pairs such that pairs by the same authors have high scores and those by different authors have lower scores.  

Below we define the components of our primary models.  We also note alternative model choices for the sub-networks, which we examine after the main results.

\subsection{Sub-networks}

Like \mycitet{koch2015siamese}, we used CNNs here, in line with the observation of \mycitet{kim:2014:EMNLP} that CNNs are good at text classification.  Our primary sub-network architecture is similar to that of \mycitet{ruder2016character}, a high-performing CNN classification approach to authorship attribution.  The input for our primary model is character-level: \mycitet{ruder2016character} found that character-level input almost always worked best, and the representation is also character-level in \mycitet{koppel2011authorship}, in line with the observations of \mycitet{keselj-etal:2003:PACLING} about stylistic authorship classification.
Each sub-network consists of an embedding layer, four convolutional layers, and a dense layer. 
The activation functions are tanh for convolutional layers and sigmoid for dense layers.

We do also examine the effect of different choices here: choosing word-level input instead of character-level input, and LSTMs instead of CNNs.
For the LSTM alternatives, we look at both unidirectional (left-to-right) and bidirectional.


\subsection{Energy functions}

\mycitet{koch2015siamese} considered both the $L_1$ and $L_2$ distances between the outputs of the final layers of their sub-networks (vectors $v_1, v_2$ in our Fig~\ref{fig:siamese-net}), and found that $L_1$ worked better for their image matching task.  Adapting their notation, we use this same function for our distance calculation:

\[
\mathbf{p} = \sigma(\sum_j \alpha_j \, | v_1^{(j)} - v_2^{(j)} \, |)
\]

\noindent
where $v_i$ is the output of the final layer of sub-network $i$ (in our case, the dense layer after the convolutional layers) and $v_i^{(j)}$ the $j$th element of it;
$\alpha_j$ the additional parameters that are learned
by the model during training, weighting the importance
of the component-wise distance; and $\sigma(\cdot)$ the sigmoid activation function.  This defines the final fully-connected layer for the network which joins the two Siamese components.  When applied to our CNN sub-networks described above, we refer to the architecture as \SiamL.

We also observe, however, that in text-related tasks cosine similarity is commonly used: this is the measure used in \mycitet{koppel2011authorship} and many text-based Siamese or DSSM models, as discussed in \S\ref{sec:rel-work}.  We therefore introduce a variant of \SiamL where the distance calculation is the complement of the cosine similarity between $v_1$ and $v_2$, similar to \mycitet{rodrigues2017ways}.  As this is a scalar quantity, there is no final dense layer.  The energy function is then:

\[
\mathbf{p} = \cossim(v_1, v_2)
\]

We refer to this as \SiamCOS.



\section{Experimental Setup}
\label{sec:exper}

\subsection{Evaluation Framework}
\label{sec:eval-framework}

\subsubsection{Known Author vs One-Shot}

We consider two types of evaluation.  The first is the one-shot evaluation of \mycitet{koch2015siamese}. Here the set of authors in the test set is \textit{disjoint} with respect to the authors in the training set.  Closed-set classification approaches do not apply here, as there is no way to build a model of a previously unseen author.  Open-set approaches will only work to the extent that they embody general notions of stylistic similarity between authors.  We refer to this as the \OneShot setup.  In \OneShot, the training set consists of 2/3 of the authors, as described below.

The second type of evaluation is common in authorship attribution: while the texts in the training and test sets are different, the same set of authors is represented in both.  We refer to this as the \KnownAuth setup.  Classification approaches are applicable here, as well as similarity; for the similarity approaches, what they embody could involve both properties of specific authors and general models of authorial similarity.  In \KnownAuth, the training set consists of 3/4 of the texts written by each author.

\subsubsection{Verification vs N-way}

As in \mycitet{koch2015siamese}, we begin with the task of \textit{verification}: Are two texts by the same author?  We use this solely to investigate how our Siamese models perform on their fundamental task of scoring similar authors high and different authors low. 

The main task, also framed as in \mycitet{koch2015siamese}, is \textit{N-way} evaluation: Given a text $T$ by author $\mathcal{A}$, select the text out of $N$ candidates that is also by $\mathcal{A}$; there will be exactly one by $\mathcal{A}$ among the $N$.

The N-way evaluation applies to both \KnownAuth and \OneShot frameworks.  The similarity approaches choose the candidate from among the $N$ that has the highest similarity score to $T$.  For the classification approach in \KnownAuth (\S\ref{sec:baselines}), the candidate that is chosen is the author with the highest network prediction among the $N$.

\subsection{Data}
\label{sec:data-set}

\subsubsection{Datasets}

There are several datasets previously employed for author identification, including various PAN datasets.
While the PAN datasets released up to 2019 
have been used by a number of authors, they are small for our many-author setup (the largest has 180 authors) and too small to train a deep learning model.
For benchmarking the verification task on small data, we do use the PAN 2015 data, which consists of 4 languages (English, Dutch, Spanish and Greek). Each language includes 100 instances for which one unknown piece and one to five known pieces are given \citep{stamatatos:2015b}. The number of authors is unknown, and the genre of text varies.
We will refer to this dataset as \PanF.

Some other large data sets are the Enron emails corpus; a set of IMDB reviews; and the Blog Authorship Corpus \cite{schler-etal:2006}, a large sample of personal blogs collected from \texttt{blogger.com}.\footnote{\url{http://u.cs.biu.ac.il/~koppel/BlogCorpus.htm}}
We use the last of these as it includes a sufficiently large number of authors for our purposes: 
we extracted a subset of 1950 authors that contains all blogs with at least 1500 words, and retain as the text the first 1000 words.  The average number of samples per author is 2.83, and the average vocabulary size under character-level tokenization is 270.
We refer to this dataset as \Blog.  

In addition, we use a more recent dataset put together by \cite{fernandes-etal:2019:POST}.
Like the PAN 2018 attribution task, it consists of fanfiction; we choose this dataset as it has more authors.
It was collected from \texttt{fanfiction.net} from the five most popular fandoms (``Harry Potter", ``Hunger Games", ``Lord of the Rings", ``Percy Jackson and the Olympians" and ``Twilight").\footnote{Available at \url{https://github.com/ChakavehSaedi/Siamese-Author-Identification}.}
We observe that having authors writing on similar topics (within a small number of ``fandoms'') means that methods cannot rely on topic cues.
From this we have put together 4 subsets of varying numbers of randomly chosen authors (100, 1K, 5K and 10K).  
Each text consists of 2000 words.  The average number of samples per author is 2.1, and the average vocabulary size under character-level tokenization is 365.
We will refer to these datasets as \FF-$n$, where $n$ is the number of authors.

We also use the dataset from the author verification task of PAN 2020.
There are two versions of the dataset, large and small, where the latter is a subset of the former. Like many of the PAN participants, we train our network on the smaller version, although we also use the larger one as described below in \S\ref{sec:training-data}.
The small training data for author verification\footnote{Restricted access is provided at \url{https://zenodo.org/record/3724096\#.YBC0PugzZPY}} includes over $50K$ pairs extracted from $1600$ fandoms written by $6400$ authors in positive pairs (i.e. pairs written by the same author) and $48500$ authors in negative pairs (i.e. pairs written by different authors). Each text piece is of approximate length of $21000$ characters and average word count of $4875$. 
We use $90\%$ of the data for training and $10\%$ for validation, and will refer to this dataset as \PanT.  (The testset used in the PAN 2020 task is not publicly available.)
As this dataset is for the verification task, we need to transform it to our N-way set-up as described in \S\ref{sec:training-data}.

For all datasets, we did not employ any specific pre-processing such as lemmatization or lower-casing, nor did we replace digits, letters or punctuation, as these can be indicators of authorship. PAN 2020, however, applied white space and punctuation normalization when preparing the data \mycitep{kestemont2019overview}.

\subsubsection{Training Data}
\label{sec:training-data}

To produce a reasonable number of samples, we divide each text into 8 pieces. In order to generate same/different author pairs for training the Siamese networks, the pieces are divided into 4 chunks, which are then paired, as follows. 

\begin{figure}
  \centering
  \includegraphics[width=70mm]{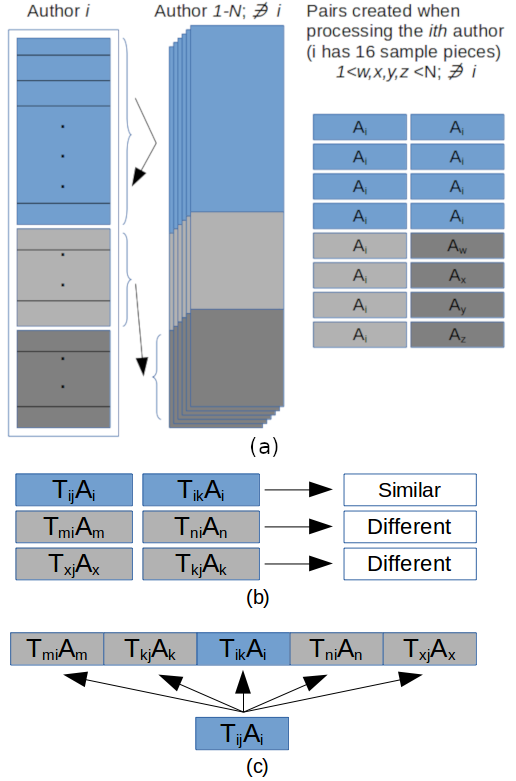}
  \vspace{3\baselineskip}
  \caption{Schematic view of (a) how pieces are randomly paired to create similar and different entries for train and test. (b) Test-set data for binary classification (author verification). (c) One set of 5-way \OneShot task randomly selected from the testset, where $T_{ij}A_{i}$ stands for the $j$th piece written by the $i$th author.}
  \label{fig:schema}
\end{figure}

Dividing each text into 8 pieces, if there are $N$ authors and $M$ documents per author, this gives $8\times{N}\times{M}$ pieces. In order to generate same/different author pairs for training the Siamese networks, the pieces are divided into 4 chunks. Pieces included in the first two chunks for an author $\mathcal{A}$ (colored blue in Figure~\ref{fig:schema}(a)) are randomly paired together to create same-author pairs. For different-author pairs, pieces in the third chunk (light gray in Figure~\ref{fig:schema}(a)) for author $\mathcal{A}$ are paired with pieces in the fourth chunk (dark gray in Figure~\ref{fig:schema}(a)) for some other author $\mathcal{B}$; both selections (of author $\mathcal{B}$ and of sample piece) are randomized.  In this way, we make sure none of the samples forming similar pairs are used more than once. Figure~\ref{fig:schema}(b) illustrates a schematic train/test-set where $T_{ij}A_{i}$ stands for the $j$th piece written by the $i$th author.

Table \ref{tab:pair_no} shows the number of pairs making up these datasets.

\begin{table}
\caption{Number of pairs included in the train and test sets for the Siamese network. $\star$ Extracted from PAN 2020 large, only containing authors that exist in the training set.  $\diamond$ Extracted from PAN 2020 large, authors are disjoint from the training set.}
  \vspace{0.5\baselineskip}
\centering
\begin{tabular}{l|rr|rr} 
\toprule
{} & \multicolumn{2}{|c|}{\KnownAuth} & \multicolumn{2}{c}{\OneShot}\\
\midrule
dataset  &Train   &Test    &Train     &Test    \\
\midrule
\FFa     &600     &200     &520       &280    \\
\FFb     &6044    &2010    &5350      &2700   \\
\FFc     &32350   &10658   &29650     &13700  \\
\FFd     &69710   &22818   &64670     &29120  \\
\midrule
\Blog    &13410   &2580    &12730     &6380   \\         
\midrule
\PanF    &various  &various   &various    &various  \\
\PanT    &275565  &12000$\star$   &275565    &12000$\diamond$  \\
\bottomrule
\end{tabular}
\label{tab:pair_no}
\end{table}

The final train and test sets are balanced in the number of similar and different pairs.  We keep $10\%$ of the training set for validation data. 

\paragraph{PAN Datasets}
\label{pan2015Pairing}
The text chunks in the PAN 2015 dataset do not come in pairs. We employed four different strategies for pairing which enables us to study the effect of text size as well as order of chunks:
\begin{enumerate}[(A)]
  \item Every unknown entry is paired with each of the elements in the known set. The whole text from both unknown and known samples is used as one piece.
  \item Every unknown entry is paired with each of the elements in the known set. Only the beginning part from both unknown and known samples is used, the size is equal to the size of the shortest sample in the language cases.
  \item Each sample in the language cases is divided into three equal pieces. First, second and third pieces from the unknown sample are paired with first, second and third pieces in the known samples respectively.
  \item Each sample in the language is divided into three equal pieces. First, second and third pieces from the unknown sample are randomly paired with first, second and third pieces in each known sample.
\end{enumerate}

The \PanT dataset provided positive and negative pairs set up for the verification task.  To construct data for our experiments:
\begin{itemize}
    \item For the \OneShot setup, we extracted $12,000$
pairs from the PAN 2020 large training subset, such that no authors were in our training set.
    This testset includes $13,000$ authors, making it our largest.  We constructed N-way sets from this.
    \item For the \KnownAuth setup, we extracted $12,000$ pairs from the PAN 2020 large training subset, and similarly constructed N-way sets from this.
    This testset includes $12,689$ authors, all of which are included in \PanT training set.
\end{itemize}


\subsubsection{Test Data and Evaluation Metric}
\label{sec:test-eval}

For N-way evaluation, we randomly create 500
sets of N-way authors from the appropriate test set (\KnownAuth or \OneShot), and we calculate the accuracy in predicting the correct author.  Final results are based on the average of three runs of different sets of 500.  For verification, we report results on all elements of the test set.



\paragraph{Additional PAN 2020 Evaluation}
In addition to our core evaluation above, for \PanT we calculate the evaluation metrics used in the PAN 2020 verification task, using the provided code.\footnote{\url{https://github.com/pan-webis-de/pan-code/tree/master/clef20/authorship-verification}}. The four metrics are:
\begin{itemize}
\item
\textbf{AUC} (sometimes referred to as ROC-AUC), which calculates the area under curve score.
\item 
\textbf{F1}, the conventional F1 score calculated using precision and recall.
\item
\textbf{F0.5u}, a new measure emphasising correct same-author predictions.
\item 
\textbf{C@1}, which is very similar to the conventional F1 score. C@1 rewards the model if hard problems are not answered (i.e. a score of $0.5$ is considered to indicate ``I don't know".).
\end{itemize}

\subsection{Baselines}
\label{sec:baselines}

\subsubsection{Similarity}

As noted in \S\ref{sec:introduction}, the most prominent authorship conventional similarity-based method is by \cite{koppel2011authorship}.
To our knowledge, this is the only available method that can be used as is in our one-shot experimental setup.\footnote{The other main method of \mycitet{abbasi-chen:2008:TOIS} appears not to have an available implementation or sufficient detail for reimplementation.}
We used as a starting point 
code from a reproducibility study
\mycitep{potthast-etal:2016:ECIR}; we reimplemented it to improve performance.  We refer to this as \Koppel.

\subsubsection{Classification}

As noted in \S\ref{sec:model}, the sub-networks in our Siamese architecture are similar to the high-performing method of \mycitet{ruder2016character} (see \S\ref{sec:rel-work}).  We use an individual sub-network as our classification architecture.  We refer to this as \CNN.

As another baseline, we consider the type of approach based on language model pretraining that has recently come to dominate performance in many NLP tasks.  In these, pretrained language representations can be used either as additional features in a task-specific architecture (e.g. ELMo: \mycitet{peters-etal:2018:NAACL}) or via transfer learning and the fine-tuning of parameters for a specific task (e.g. GPT: \mycitet{radford-etal:2018}).  BERT \mycitep{devlin-etal:2019:NAACL} is an approach that when it was introduced produced state-of-the-art performance on a range of NLP tasks set up as the GLUE benchmark\footnote{\url{https://gluebenchmark.com/}} \mycitep{wang-etal:2018:EMNLP}: it gave the best performance on all tasks in this suite, including sentiment classification, prediction of grammatical acceptability, textual similarity, paraphrase, and natural language inference; improvements on many of the tasks were quite large with respect to previous state of the art.  Later analysis \mycitep{tenney-etal:2019:ACL} has shown that BERT can perform across levels of linguistic analysis, from low (e.g. part-of-speech tagging) to high (e.g. semantic roles).

We therefore use BERT fine-tuned for our classification task as our second baseline.  We do this by feeding the output of BERT to a dense layer, and carrying out a small amount of extra training.


\subsection{Implementation Details}

\subsubsection{Siamese Networks}

In terms of structure, each sub-network consists of an embedding layer, four convolutional layers, and a dense layer (resp. Emb, Conv$n$, D in Table~\ref{tab:net-par}).

\begin{table}
\centering
\caption{Network parameters}
  \vspace{0.5\baselineskip}
\begin{tabular}{lrrrrrr}
\toprule
Type   &Emb    &Conv1  &Conv2  &Conv3  &Conv4  &D      \\
\midrule
Size   &300    &350   &300   &250   &250   &400     \\
Filter Size &    &1     &2     &3     &3     &       \\            
\bottomrule
\end{tabular}
\label{tab:net-par}
\end{table}

The Siamese networks are trained on the verification task, for at most 25 epochs. All initializations are random, and training is restarted if after the $10$th epoch the verification accuracy is smaller than $0.55$.

The epoch we select for the final result is the one with best verification accuracy on the validation set.

In terms of hyper-parameters, we use a learning rate of 0.0005, Adam as optimizer, and batch size of 25.

As noted above, for the LSTM alternatives to the main models, we look at both unidirectional (left-to-right) and bidirectional.  In these, we use a hidden layer with 200 nodes.
For the LSTM variants, hyperparameters have the same settings.


\subsubsection{Koppel}

\Koppel has few parameters.  The maximum number of character 4-grams is set to 20,000 as in the replication code; the actual number of character 4-grams in our data is always lower than this.  The replication code samples 50\% of the features, and repeats this 100 times, which \mycitet{koppel2011authorship} found to produce good results.  The replication code also by default uses the Ruzicka metric rather than cosine similarity (which we also found to perform better).  There is an additional parameter, a threshold for a `don't know' option; we always make a choice, and so set this threshold to be 0.

We reimplemented the replication code to be more efficient, in order to run on larger numbers of authors: the replication code did not, for example, have efficient implementations of vector arithmetic.  We verified that the replication code and our reimplementation performed the same on the PAN 2011 and 2012 and \FFa datasets.  Results in the paper are all from our reimplementation.

\subsubsection{CNN Classification}

The CNN classification model is trained for at most 150 epochs, and the epoch with the best validation accuracy across all classes is selected.

\subsubsection{BERT Classification}

To fine-tune BERT for authorship attribution, we trained for 3 epochs, as did \cite{devlin-etal:2019:NAACL} for all GLUE benchmark tasks.  BERT takes as input sentences, so we segmented our input at the period character.  (Other segmentations produced similar results, although they declined more steeply for larger $N$.)


\section{Experimental Results}
\label{sec:results}

\begin{table}
  \vspace{1.5\baselineskip}
    \caption{Verification: accuracy on \FF, \Blog and PAN datasets.}
  \vspace{0.5\baselineskip}

    \centering
    \begin{tabular}{l|cc} 
    \toprule
    DB     & \SiamL  & \SiamCOS     \\
    \midrule
    \FFa   &0.723    &0.674    \\
    \FFb   &0.980    &0.978    \\
    \FFc   &0.986    &0.983    \\
    \FFd   &0.982    &0.958    \\
    \midrule
    \Blog  &0.963    &0.981    \\
    \midrule
    \PanF  &0.654    &0.724    \\
    \PanT  &0.782    &0.883    \\
    \bottomrule
    \end{tabular}
    \label{tab:ver}
    \end{table}

\subsection{Verification}
\label{sec:results-verif}

\paragraph{Core Results}
Table~\ref{tab:ver} gives the results for author verification for our two Siamese variants.  (For PAN 2015, this is the best result for the English subset of the data.)
As expected, accuracy improves with more training data: it starts very low when there are only 100 authors to learn a notion of similarity from, increasing rapidly when there are 1000 authors to 0.980 for \SiamL and 0.978 for \SiamCOS;  there is no improvement for 10000 authors.  The scores on \Blog are very close to what we have for the similarly sized \FF datasets.  For the PAN datasets, again the larger has higher scores, although with PAN 2020 being the largest dataset and with the longest texts, it may seem surprising that the results are not higher.  To handle the larger text sizes --- text chunks in \PanT are approximately $8$ times larger than for \FF --- the network needs many more epochs to converge.  The task is nevertheless still more complex, even with additional training, as there are more fandoms in \PanT compared to \FF ($1600$ vs $5$, respectively). This allows a more strict setup making the problem more complex: in \PanT, no positive pairs exist where the texts are from same fandom.

In terms of our energy function, for all of the results, there is no clear preference regarding \SiamL versus \SiamCOS.

\paragraph{Additional Results: PAN 2015}
As noted in \S\ref{sec:training-data}, the format of \PanF allowed us to try different pairing strategies to see which worked best as training data for a Siamese network for verification.  Results of all variants are in Table~\ref{tab:PAN15-eng}.  There are clear differences: in the PAN 2015 setup where there are many known author texts for each unknown author snippet, pairing this snippet up multiple times provides the highest results.

\begin{table}[!ht]
\centering
\begin{tabular}{p{3cm}p{1.5cm}p{1.5cm}p{1.5cm}p{1.5cm}} 
\toprule
Input format  &A     &B      &C      &D\\
\midrule
\SiamCOS     &0.724  &0.690  &0.563  &0.605  \\
\SiamL       &0.654  &0.680  &0.650  &0.633  \\
\bottomrule
\end{tabular}
\caption{Best Results on Author Verification for English subset of \PanF.}
\label{tab:PAN15-eng}
\end{table}

Results for the other languages in the dataset, using the setting under which we got the highest accuracy for the English dataset (A), are shown in Table~\ref{tab:PAN15-dsg}. 
Here the datasets are small enough that results are sometimes not much over random chance (0.5 in our verification setup), reinforcing that larger datasets are necessary for these deep Siamese techniques.

\begin{table}[!ht]
\centering
\begin{tabular}{p{3cm}p{1.5cm}p{1.5cm}p{1.5cm}} 
\toprule
Input language  &Dutch    &Spanish     &Greek\\
\midrule
\SiamCOS        &0.582    &0.762       &0.593  \\
\SiamL          &0.563    &0.650       &0.550   \\
\bottomrule
\end{tabular}
\caption{Author Verification accuracy for Dutch, Spanish and Greek data set of \PanF.}
\label{tab:PAN15-dsg}
\end{table}

\paragraph{Additional Results: PAN 2020}
As noted in \S\ref{sec:data-set}, the official testset for PAN 2020 is not available.
Results for \PanT are provided in Table~\ref{tab:PAN20}, under
our setup where we use $90\%$ of \PanT for training and the rest for validation.  In addition to the usual accuracy, 
we also calculated the four PAN 2020 metrics described in \S\ref{sec:test-eval}.

While this cannot be directly compared with results in PAN 2020, if our verification set were to have the same distributional properties and other characteristics as the official test set, \SiamCOS would rank second among competition systems, according to the data provided in \cite{Kestemont2020OverviewOT}.

\begin{table}
\centering
\begin{tabular}{l|cccccc}
\toprule
Model        &Epoch &Accuracy  &ROC-AUC  &F1    &C@1   &F0.5u  \\
\midrule
\SiamCOS     &29    &0.883     &0.954    &0.878 &0.883 &0.886 \\
\SiamL       &39    &0.782     &0.849    &0.778 &0.782 &0.783 \\
\bottomrule
\end{tabular}
\caption{Author verification accuracy, AUC, F1, C@1, and F0.5u scores on \PanT using the two variants of the Siamese model.}
\label{tab:PAN20}
\end{table}

As observed in \S\ref{sec:training-data}, \PanT text chunks are up to $8$ times larger than the chunks used for training the Siamese model on the \FF datasets. This provides us with an opportunity to analyse the effect of text size on accuracy on both Siamese variants.

\begin{figure}[!ht]
\centering
\includegraphics[width=95mm]{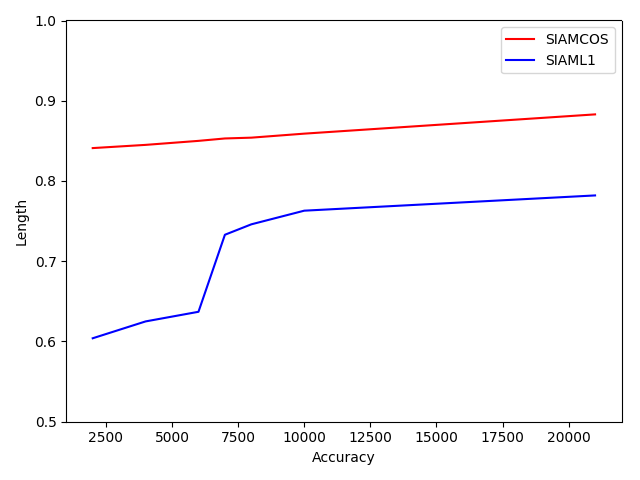}
\caption{Effect of text length on the accuracy of \SiamCOS and \SiamL}
\label{fig:txtLenEffect}
\end{figure}

Figure \ref{fig:txtLenEffect} shows that, as expected, the larger the text chunk, the more authorial features there are to rely on and the higher the accuracy. \SiamCOS is consistently above \SiamL for all text chunk sizes on this dataset.
Interestingly, there is a steady linear relation between \SiamCOS's accuracy and the text chunk size, while \SiamL shows a jump when moving from size $6000$ to $7000$ characters.

\begin{table}
\renewcommand{\arraystretch}{0.97}
\centering
\caption{Results under the \OneShot scenario on \FFb, \FFc, \Blog and \PanT: N-way classification accuracy.}
\small{
\begin{tabular}{l|r|ccc} 
\toprule
& $N$      &\Koppel & \SiamL  & \SiamCOS\\
\midrule
\multirow{6}{*}{\FFb} 
& 2      &0.783      &0.957  &0.990  \\
& 3      &0.697      &0.893  &0.987  \\
& 5      &0.533      &0.847  &0.950  \\
& 10     &0.440      &0.790  &0.943  \\
& 20     &0.340      &0.640  &0.907  \\
& 50     &0.247      &0.553  &0.870  \\
& 100    &0.160      &0.440  &0.753  \\
& 300    &0.117      &0.373  &0.547  \\

\midrule
\multirow{7}{*}{\FFc} 
& 2     &0.757      &0.967  &0.997   \\
& 3     &0.620      &0.923  &0.999   \\
& 5     &0.510      &0.883  &0.990   \\
& 10    &0.457      &0.803  &0.997   \\
& 20    &0.317      &0.693  &0.987   \\
& 50    &0.274      &0.570  &0.957   \\
& 100   &0.163      &0.513  &0.930   \\
& 500   &0.123      &0.377  &0.780   \\

\midrule
\multirow{7}{*}{\Blog} 
& 2      &0.833      &0.900  &0.983    \\
& 3      &0.707      &0.837  &0.973    \\
& 5      &0.617      &0.713  &0.977    \\ 
& 10     &0.433      &0.577  &0.947    \\
& 20     &0.360      &0.490  &0.880    \\
& 50     &0.273      &0.383  &0.853    \\
& 100    &0.158      &0.347  &0.813    \\
& 500    &0.136      &0.270  &0.520    \\

\midrule
\multirow{7}{*}{\PanT} 
& 2      &0.870      &0.820  &0.930    \\
& 3      &0.730      &0.713  &0.877    \\
& 5      &0.610      &0.637  &0.850    \\ 
& 10     &0.640      &0.503  &0.723    \\
& 20     &0.420      &0.383  &0.687    \\
& 50     &0.390      &0.223  &0.567    \\
& 100    &0.270      &0.167  &0.480    \\
\bottomrule
\end{tabular}
}
\label{tab:os-all}
\renewcommand{\arraystretch}{1.0}
\end{table}

\begin{figure}
\centering
\includegraphics[width=75mm]{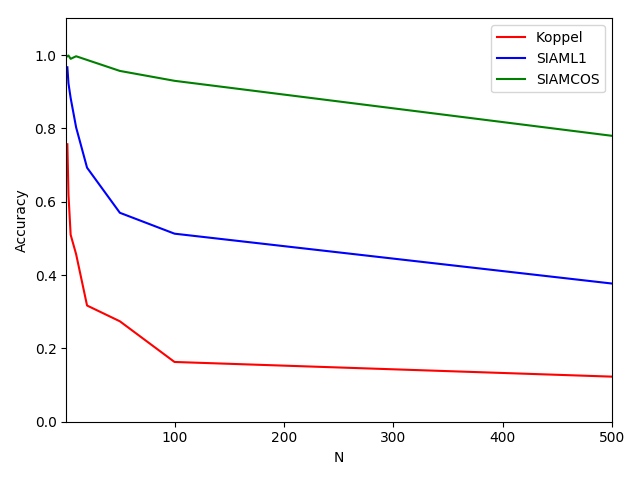}
\vspace{\baselineskip}

\caption{Results on \FFc under the \OneShot scenario: accuracy under N-way classification.}
\label{fig:N-way_chart}
\end{figure}

\subsection{N-Way One-Shot}

\subsubsection{Core Model}
\label{sec:results-core}

The verification results above indicate that 100 authors do not provide enough data for the Siamese networks to train to a high level of performance, and that results for 10000 authors as roughly the same as for 5000 authors.  For the N-way one-shot scenario, then, Table~\ref{tab:os-all} presents results for \FFb, \FFc, \Blog and \PanT.

We make the following observations:

\begin{itemize}
\item 
All results are much higher than chance ($= 1/N$), and naturally degrade as $N$ increases. 

\item
\SiamCOS has the best results for all datasets and for all $N$ over those datasets.
\SiamL, while not as good as \SiamCOS, is almost always better than \Koppel; the exception is \PanT.

\item
Performance on the datasets for \SiamCOS and \SiamL is generally similar on each dataset for corresponding values of $N$.  \PanT results are a little lower for all $N$ (e.g. for \SiamCOS, for $N = 2$: 0.930 vs 0.983--0.990; for $N = 100$: 0.480 vs 0.753--0.930).  There are two factors here acting in opposite directions: making the task more difficult than on other datasets is the very large number of authors, but making it easier is the larger text size.  We explore the effect of text size in \S\ref{sec:results-alts}.

\item
\Koppel performs relatively better on \Blog (which was its original test corpus) than on the \FF corpora, that is, comparing like $N$s against the performance of \Koppel on the other corpora; for example, it scores 0.617 accuracy on \Blog for 5-way comparison on this dataset of 2000 authors, versus 0.533 for 5-way comparison on \FFb. (However, its performance is still lower than \SiamL.)  \Koppel similarly performs relatively better on \PanT, where it also beats \SiamL.
In the case of \Blog, this may because of the observation in \mycitet{koppel2011authorship} that the method could rely on topic clues, which they see as reflective of real-world use, as blog posts by a given author commonly share topics that other authors may not discuss; in the \FF datasets, the topics (fandoms) are shared across many authors.  This is not likely to be the explanation for \PanT, where the dataset pairs do not share a topic (fandom); rather, it may be because the texts are larger.  \SiamCOS seems relatively invariant to either of these effects.


\end{itemize}

\subsubsection{Model and Training Alternatives}
\label{sec:results-alts}

\paragraph{Character versus Word}
In addition to the architectural choices for our primary model as described in \S\ref{sec:model}, we also tried word-level inputs, and these as expected performed consistently worse, indicating stylistic features can be better identified through characters. Table \ref{tab:word} shows a comparison between word- and character-level inputs on \FFb under \SiamCOS and \SiamL.
It is apparent that the difference is large and gets dramatically larger as $N$ increases.  In the pre-deep-learning era, \mycitet{keselj-etal:2003:PACLING} argued that character-level representations better capture stylistic characteristics for authorship; this is supported by these results, and indicates that it continues to be true for deep learning representations.

\begin{table}
\centering
\caption{N-way accuracy under word- and character-level inputs on \FFb under the \OneShot scenario}
  \vspace{0.5\baselineskip}

\begin{tabular}{lrrrrrr}
\toprule
N   &Model      &2      &3      &5      &10     &100   \\
\midrule
\multirow{2}{*}{Char}
&\SiamCOS    &0.990 	&0.987 	&0.950 	&0.943 	&0.753     \\
&\SiamL      &0.957 	&0.893 	&0.847 	&0.790 	&0.440     \\
\midrule
\multirow{2}{*}{Word}
&\SiamCOS    &0.710 	&0.523 	&0.420 	&0.240 	&0.047     \\
&\SiamL      &0.597 	&0.450 	&0.32 	&0.130 	&0.013     \\

\bottomrule
\end{tabular}
\label{tab:word}
\end{table}



\paragraph{CNN versus LSTM}
Another alternative discussed in \S\ref{sec:model} was to use LSTMs for sub-networks; results, again on \FFb, are in Table~\ref{tab:LSTM}.  Here also the results of the LSTMS are substantially worse than the core \SiamCOS model using a CNN, and only the biLSTM for the largest $N$ is better than the core \SiamL model.  In addition, in terms of practical use for these larger $N$, the LSTM-based networks are much slower to train, taking orders of magnitude longer: for the 100-way setup, one epoch took around a day.  They are clearly infeasible for this kind of authorship identification setup.

\begin{table}
\centering

  \vspace{1.5\baselineskip}
\caption{N-way accuracy using LSTMs vs CNNs on \FFb under the \OneShot scenario}
  \vspace{0.5\baselineskip}

\begin{tabular}{lrrrr}
\toprule
N             &2        &3      &10        &100  \\
\midrule
CNN (\SiamCOS)       &0.990 	&0.987 	&0.943     &0.753     \\
CNN (\SiamL)         &0.957 	&0.893 	&0.790     &0.440     \\
LSTM (LtR)    &0.484 	&0.332 	&0.116 	   &0.04     \\
LSTM (bi-dir) &0.898    &0.878  &0.780     &0.526\\
\bottomrule
\end{tabular}
\label{tab:LSTM}
\end{table}




We also considered both $L_2$ as a variant of \SiamL and the Ruzicka or minmax metric as a variant of \SiamCOS, as this latter has been found to be an improvement of \mycitet{koppel2011authorship} by \mycitet{koppel-winter:2014:JASIST}.  Again, results were consistently poorer and we do not present them.

\begin{table}[]
    \centering
    \begin{tabular}{r|rr|rr}
    \toprule
    {} & \multicolumn{2}{|c|}{\PanT: 2700 chars} & \multicolumn{2}{c}{\PanT: full}\\
    \midrule
    N        &\SiamL   &\SiamCOS   &\SiamL   &\SiamCOS \\
    \midrule
    2        &0.770    
    & 0.830     &0.820  &0.930    \\
    3        &0.530    
    & 0.760     &0.713  &0.877    \\
    5        &0.390    
    &0.620      &0.637  &0.850    \\ 
    10       &0.220    
    &0.470      &0.503  &0.723    \\
    20       &0.190    
    &0.410      &0.383  &0.687    \\
    50       &0.050    
    &0.290      &0.223  &0.567    \\
    100      &0.050    
    &0.160      &0.167  &0.480    \\
    \bottomrule
    \end{tabular}
    \caption{$N$-way classification accuracy under \OneShot scenario on chunks of maximum 2700 character in \PanT pairs, compared to scores for full-length texts from Table~\ref{tab:os-all}, using \SiamL and \SiamCOS.}
    \label{tab:PAN20-shorter-texts}
\end{table}

\paragraph{Training Pair Size}
As noted in \S\ref{sec:data-set}, the individual texts of \PanT are much larger than for the other datasets.  As in \S\ref{sec:results-verif}, we therefore use this dataset to examine the effect of text size, here under the N-way setup.  In Table~\ref{tab:PAN20-shorter-texts} we compare the results of training the core model on texts that have the same average size as the \FFb dataset --- 2700 characters --- with the results from Table~\ref{tab:os-all} on the full texts.  For both \SiamL and \SiamCOS, results are quite dramatically lower relative to the full-sized texts, and also relative to corresponding values of $N$ for the other datasets.  As discussed in \S\ref{sec:results-core}, the very large number of authors produces this much lower result.

\subsection{N-Way Known Author}

Table~\ref{tab:ka-1K} shows the results for \FF under the \KnownAuth scenario: we chose the smallest of the three datasets from Table~\ref{tab:os-all}, \FFb, so that the classification approach would be competitive.  

\begin{itemize}
    \item 
For the smallest case, of $N=2$, CNN classification does better than the traditional \Koppel similarity, although it degrades much more quickly as $N$ grows: this conforms to the general belief that conventional similarity methods work better for large numbers of authors.  

\item
BERT follows a similar pattern.  It starts slightly lower than CNN --- as it uses word-level (or at least word piece-level) representations, this is not unexpected, in spite of its strong performance on other tasks --- but degrades more slowly.

\item
Our new methods, \SiamL and \SiamCOS, behave similarly to the \OneShot scenario.  \SiamCOS starts as the highest at $N = 2$, and stays the best until $N = 500$, at which point \Koppel is slightly higher.

\item
Comparing Table~\ref{tab:os-all} and Table~\ref{tab:ka-1K}, it can be seen that the Siamese scores are uniformly lower for equivalent $N$.  This is because the network receives as input only 3/4 of the data per author (with 1/4 held out for \KnownAuth testing).  We would expect that with quantities of training text per author that are similar to the \OneShot scenario, we would see the same higher levels of accuracy for the Siamese methods.
\end{itemize}

\begin{table}
    \centering
    \caption{Results under the \KnownAuth scenario on \FFb: N-way classification accuracy.}

\vspace{0.5\baselineskip}

    \begin{tabular}{r|rrrrr}
    \toprule
    $N$    &\CNN   &BERT   &\Koppel  &\SiamL    &\SiamCOS \\
    \midrule
    2      &0.855  &0.828  &0.873    &0.850     &0.960    \\
    3      &0.728  &0.848  &0.713    &0.730     &0.893    \\
    5      &0.601  &0.627  &0.607    &0.657     &0.870    \\
    10     &0.488  &0.516  &0.540    &0.490     &0.757    \\
    50     &0.242  &0.262  &0.370    &0.297     &0.413    \\
    100    &0.172  &0.214  &0.230    &0.200     &0.293    \\
    500    &0.083  &0.100  &0.153    &0.137     &0.143    \\
    \bottomrule
    \end{tabular}
    \label{tab:ka-1K}

\end{table}





Table~\ref{tab:Pan20Nway} gives results for \PanT.  For this dataset we only consider \Koppel, \SiamL and \SiamCOS, given the very large number of authors.  As elsewhere, \SiamCOS is the best; and also as observed above for \FFb, results on \KnownAuth are lower than under \OneShot (Table~\ref{tab:os-all}) for corresponding values of $N$, supporting our suggestion that it is the smaller amount of training data per pair in the \KnownAuth setup.


\begin{table}[!ht]
    \centering
    \begin{tabular}{r|rrr}
    \toprule
    $N$        & \Koppel    &\SiamL     &\SiamCOS     \\
    \midrule
    2          &0.930       &0.940      &0.960        \\
    3          &0.820      &0.760      &0.920         \\
    5          &0.700       &0.650      &0.890        \\
    10         &0.600       &0.550      &0.820        \\
    20         &0.560       &0.350      &0.710        \\
    50         &0.458       &0.200      &0.630        \\
    100        &0.304       &0.130      &0.560        \\
    \bottomrule
    \end{tabular}
    \caption{$N$-way classification accuracy on \PanT under \KnownAuth scenario.}
    \label{tab:Pan20Nway}
\end{table}

\section{Conclusion}

In this work we have presented an investigation of the application of a Siamese network architecture to large-scale stylistic author attribution.
Our system learns a general notion of authorship, strongly outperforming the key similarity-based method in one-shot N-way evaluation, and also performing well in a known-author context.  While there is no clear difference between the $L_1$ metric and cosine similarity in the verification task, the latter is substantially better in a task of choosing among $N$ authors, both in open-set and closed-set contexts.
We also find that CNNs in general perform well in terms of the architectural structure of the Siamese subnetworks, and that for large numbers of authors LSTMs take infeasibly long to train.

There are two key directions we are exploring for future work.  One is with respect to the type of metric.  While we focussed on the most commonly used, notably $L_1$ and cosine similarity, hyperbolic distance has shown to useful in recent work on preserving privacy in the face of authorship attribution attacks \mycitep{feyisetan-etal:2019:ICDM}; using such a metric in authorship attribution could similarly be useful.  The other future direction is in terms of exploring other architectures used for one-shot tasks in image processing, such as the matching networks of \mycitet{vinyals-etal:2016:NIPS}: that work, for instance, incorporates other ideas from metric learning into deep learning, and aims to improve over a regular Siamese architecture by a better alignment of training objective to the N-way classification task, and by incorporating various deep learning mechanisms such as attention.  Given the success of attention-based models across a wide range of NLP tasks \mycitep{vaswani-etal:2017:NIPS}, this could be a promising next step.




\bibliographystyle{plainnat}
\bibliography{mybibfile}

\end{document}